# Mining Education Data to Predict Student's Retention: A comparative Study

Surjeet Kumar Yadav,
Research Scholar,
Sri Venkateshwara University
J. P. Nagar, India
Surjeet_k_yadav@yahoo.co.in

Brijesh Bharadwaj,
Assistant Professor, Dept. of MCA
Dr. R. M. L. Awadh University,
Faizabad, India
wwwbkb@rediffmail.com

Saurabh Pal,
Head, Dept. of MCA
VBS Purvanchal University,
Jaunpur, India
drsaurabhpal@yahoo.co.in

*Abstract*— The main objective of higher education is to provide quality education to students. One way to achieve highest level of quality in higher education system is by discovering knowledge for prediction regarding enrolment of students in a course. This paper presents a data mining project to generate predictive models for student retention management. Given new records of incoming students, these predictive models can produce short accurate prediction lists identifying students who tend to need the support from the student retention program most. This paper examines the quality of the predictive models generated by the machine learning algorithms. The results show that some of the machines learning algorithms are able to establish effective predictive models from the existing student retention data.

*Keywords*— Data Mining, Machine Learning Algorithms, Retention Management, Predictive Models

## I. INTRODUCTION

Student retention is a challenging task in higher education [1] and it is reported that about one fourth of students dropped college after their first year [1-3]. Recent study results show that intervention programs can have significant effects on retention, especially for the first year. To effectively utilize the limited support resources for the intervention programs, it is desirable to identify in advance students who tend to need the support most. In this paper, we describe the experiments and the results from a data mining techniques for the students of MCA department to assist the student retention program on campus. The development of machine learning algorithms in recent years has enabled a large number of successful data mining projects in various application domains in science, engineering, and business [4, 5]. In our study, we apply machine learning algorithms to analyze and extract information from existing student data to establish predictive models. The predictive models are then used to identify among new incoming first year students those who are most likely to benefit from the support of the student retention program.

Prediction models that include all personal, social, psychological and other environmental variables are necessitated for the effective prediction of the retention rate of the students. The retention of students with high accuracy is beneficial for identify the students with low academic achievements initially. It is required that the identified students can be assisted more by the teacher so that their performance is improved in future.

In this connection, the objectives of the present investigation were framed so as to assist the low academic achievers in higher education and they are:

a) Generation of a data source of predictive variables.
b) Identification of different factors, which affects a student's retention rate.
c) Construction of a prediction model using different classification data mining techniques on the basis of identified predictive variables, and
d) Validation of the developed model for higher education students studying in Indian Universities or Institutions.

## II. BACKGROUND AND RELATED WORK

The most commonly cited model of retention studies is one developed by Tinto [2]. According to Tinto's Model, withdrawal process depends on how students interact with the social and academic environment of the institution.

Kember [6] describes in an Open distance learning context, researchers tend to place more emphasis on the influence of external environment, such as student's occupation and support from their family, while the concept of social integration into an Open distance learning institution's cultural fabric, is given less weight.

A number of Open Distance Learning institutions have carried out dropout studies. Some notable studies have been undertaken by the British Open University (Ashby [7]; Kennedy & Powell [8]). Different models have been used by these researchers to describe the factors found to influence student achievement, course completion rates, and withdrawal, along with the relationships between variable factors.

Pandey and Pal [9] conducted study on the student performance based by selecting 600 students from different colleges of Dr. R. M. L. Awadh University, Faizabad, India. By means of Bayes Classification on category, language and background qualification, it was found that whether new comer students will performer or not.

Hijazi and Naqvi [10] conducted as study on the student performance by selecting a sample of 300 students (225 males, 75 females). The hypothesis that was stated as "Student's attitude towards attendance in class, hours spent in study on daily basis after college, students' family income, students' mother's age and mother's education are significantly related with student performance" was framed. By means of simple linear regression analysis, it was found that the factors like mother's education and student's family income were highly correlated with the student academic performance.

Khan [11] conducted a performance study on 400 students with a main objective to establish the prognostic value of different measures of cognition, personality and demographic variables for success at higher secondary level in science stream. It was found that girls with high socio-economic status had relatively higher academic achievement in science stream and boys with low socio-economic status had relatively higher academic achievement in general.

Al-Radaideh, et al [12] applied a decision tree model to predict the final grade of students who studied the C++ course in Yarmouk University, Jordan in the year 2005. Three different classification methods namely ID3, C4.5, and the NaïveBayes were used. The outcome of their results indicated that Decision Tree model had better prediction than other models.

Pandey and Pal [13] conducted study on the student performance based by selecting 60 students from a degree college of Dr. R. M. L. Awadh University, Faizabad, India. By means of association rule they find the interestingness of student in opting class teaching language.

Ayesha, Mustafa, Sattar and Khan [14] describe the use of k-means clustering algorithm to predict student's learning activities. The information generated after the implementation of data mining technique may be helpful for instructor as well as for students.

Baradwaj and Pal [21] obtained the university students data like attendance, class test, seminar and assignment marks from the students' previous database, to predict the performance at the end of the semester.

Bray [15], in his study on private tutoring and its implications, observed that the percentage of students receiving private tutoring in India was relatively higher than in Malaysia, Singapore, Japan, China and Sri Lanka. It was also observed that there was an enhancement of academic performance with the intensity of private tutoring and this variation of intensity of private tutoring depends on the collective factor namely socio-economic conditions.

Bhardwaj and Pal [16] conducted study on the student performance based by selecting 300 students from 5 different degree college conducting BCA course of Dr. R. M. L. Awadh University, Faizabad, India. By means of Bayesian classification method on 17 attributes, it was found that the factors like students' grade in senior secondary exam, living location, medium of teaching, mother's qualification, students other habit, family annual income and student's family status were highly correlated with the student academic performance.

Yadav, Bhardwaj and Pal [17] obtained the university students data like attendance, class test, seminar and assignment marks from the students' database, to predict the performance at the end of the semester using three algorithms ID3, C4.5 and CART and shows that CART is the best algorithm for classification of data.

III. DECISION TREE INTRODUCTION

A decision tree is a flow-chart-like tree structure, where each internal node is denoted by rectangles, and leaf nodes are denoted by ovals. All internal nodes have two or more child nodes. All internal nodes contain splits, which test the value of an expression of the attributes. Arcs from an internal node to its children are labeled with distinct outcomes of the test. Each leaf node has a class label associated with it.

The decision tree classifier has two phases [4]:

i) Growth phase or Build phase.

ii) Pruning phase.

The tree is built in the first phase by recursively splitting the training set based on local optimal criteria until all or most of the records belonging to each of the partitions bearing the same class label. The tree may overfit the data.

The pruning phase handles the problem of over fitting the data in the decision tree. The prune phase generalizes the tree by removing the noise and outliers. The accuracy of the classification increases in the pruning phase.

Pruning phase accesses only the fully grown tree. The growth phase requires multiple passes over the training data. The time needed for pruning the decision tree is very less compared to build the decision tree.

A. ID3 (Iterative Dichotomise 3)

This is a decision tree algorithm introduced in 1986 by Quinlan Ross [18]. It is based on Hunts algorithm. The tree is constructed in two phases. The two phases are tree building and pruning.

ID3 uses information gain measure to choose the splitting attribute. It only accepts categorical attributes in building a tree model. It does not give accurate result when there is noise. To remove the noise pre-processing technique has to be used.

To build decision tree, information gain is calculated for each and every attribute and select the attribute with the highest information gain to designate as a root node. Label the attribute as a root node and the possible values of the attribute are represented as arcs. Then all possible outcome instances are tested to check whether they are falling under the same class or not. If all the instances are falling under the same class, the node is represented with single class name, otherwise choose the splitting attribute to classify the instances.

Continuous attributes can be handled using the ID3 algorithm by discretizing or directly, by considering the values to find the best split point by taking a threshold on the attribute values. ID3 does not support pruning.

B. C4.5

This algorithm is a successor to ID3 developed by Quinlan Ross [18]. It is also based on Hunt's algorithm. C4.5 handles both categorical and continuous attributes to build a decision tree. In order to handle continuous attributes, C4.5 splits the attribute values into two partitions based on the selected threshold such that all the values above the threshold as one child and the remaining as another child. It also handles missing attribute values. C4.5 uses Gain Ratio as an attribute selection measure to build a decision tree. It removes the biasness of information gain when there are many outcome values of an attribute.

At first, calculate the gain ratio of each attribute. The root node will be the attribute whose gain ratio is maximum.

C4.5 uses pessimistic pruning to remove unnecessary branches in the decision tree to improve the accuracy of classification.

*C. ADT (Alternating Decision Tree)*

ADTrees were introduced by Yoav Freund and Llew Mason [19]. However, the algorithm as presented had several typographical errors. Clarifications and optimizations were later presented by Bernhard Pfahringer, Geoffrey Holmes and Richard Kirkby [20].

An alternating decision tree consists of decision nodes and prediction nodes. Decision nodes specify a predicate condition. Prediction nodes contain a single number. ADTrees always have prediction nodes as both root and leaves. An instance is classified by an ADTree by following all paths for which all decision nodes are true and summing any prediction nodes that are traversed. This is different from binary classification trees such as CART (Classification and regression tree) or C4.5 in which an instance follows only one path through the tree.

The original authors list three potential levels of interpretation for the set of attributes identified by an ADTree:

- Individual nodes can be evaluated for their own predictive ability.
- Sets of nodes on the same path may be interpreted as having a joint effect
- The tree can be interpreted as a whole.

IV. DATA MINING PROCESS

Knowing the reasons for dropout of student can help the teachers and administrators to take necessary actions so that the success percentage can be improved. The data is collected from Department of MCA of V. B. S. Purvanchal University, Jaunpur, India. The raw data set is a collection of 432 records accumulated over a period of twelve years regarding the basic information of first year students and whether they continued to enroll after the first year. The MCA department has been started in the year 1997 and 12 batches have completed their study. In the raw data set, 398 of the students continued to enroll after their first year while 34 of them dropped out by the end of the first year.

*A. Data Preparations*

Data of 432 students of the Department of MCA, VBS Purvanchal University, Jaunpur is collected who get admission from 1997-2000 batch to 2009-2012 batch. The data was collected through the enrolment form filled by the student at the time of admission. The student enter their demographic data (category, gender etc), past performance data (SSC or 10th marks, HSC or 10 + 2 exam marks and Graduation Marks etc.), address and contact number.

*B. Data selection and transformation*

In this step only those fields were selected which were required for data mining. A few derived variables were selected. While some of the information for the variables was extracted from the database. All the predictor and response variables which were derived from the database are given in Table I for reference.

TABLE I: STUDENT RELATED VARIABLES

| Variables | Description | Possible Values |
|---|---|---|
| Sex | Students Sex | {Male, Female} |
| Cat | Students category | {General, OBC, SC, ST} |
| GSS | Students grade in Senior Secondary education | {O – 90% -100%, A – 80% - 89%, B – 70% - 79%, C – 60% - 69%, D – 50% - 59%, E – 40% - 49%, F - < 40% } |
| GMSS | Students grade in Math at Senior Secondary education | {O – 90% -100%, A – 80% - 89%, B – 70% - 79%, C – 60% - 69%, D – 50% - 59%, E – 40% - 49%, F - < 40%, Not Applicable} |
| GS | Graduation Stream | {BA with Math, B.A. without Math, BSc. With Math, B.Sc. without Math, B.Com, BCA, BBA, B.Tech} |
| GOG | Grade obtained in Graduation | {First $\geq$ 60% Second $\geq$ 45 & <60% Third $\geq$ 36 & < 45% } |
| MED | Medium of Teaching in Graduation | {Hindi, English, Regional} |
| CL | College Location | {Rural, Urban} |
| ATYPE | Admission Type | {UPSEE, Direct} |
| RET | Retention: Continue to enroll or not after one year | {0, 1} |

The domain values for some of the variables were defined for the present investigations are as follows:

- **Cat** – From ancient time Indians are divided in many categories. These factors play a direct and indirect role in the daily lives including the education of young people. Admission process in India also includes different percentage of seats reserved for different categories. In terms of social status, the Indian population is grouped into four categories: General, Other Backward Class (OBC), Scheduled Castes (SC) and Scheduled Tribes (ST). Possible values are General, OBC, SC and ST.

- **GSS** - Students grade in Senior Secondary education. Students who are in state board appear for five subjects each carry 100 marks. Grade are assigned to all students using following mapping O – 90% to 100%, A – 80% - 89%, B – 70% - 79%, C – 60% - 69%, D – 50% - 59%, E – 40% - 49%, and F – < 40%.

- **GMSS** – Student Grade in Mathematics at Senior Secondary education. Grade in mathematics at 10+2 level are assigned to all students using following mapping O – 90% to 100%, A – 80% - 89%, B – 70% - 79%, C – 60% - 69%, D – 50% - 59%, E – 40% - 49%, and F - < 40%. If student has not the mathematics at 10 + 2 level then assign Not-Applicable.

- **GS** – Graduation Stream. MCA admission is open for all stream students, therefore, Graduation Stream is split into following classes BA with Math, B.A. without Math, B.Sc. with Math, B.Sc. without Math, B.Com, BCA, BBA, B.Tech.

- **GOG** – Grade Obtained in Graduation. Marks/Grade obtained in graduation. It is also split into four class

values: First – ≥60% , Second – ≥45% and <60%, Third – ≥36% and < 45%.

- **Med** – This paper study covers only the degree colleges and institutions of Uttar Pradesh state of India. Here, medium of instructions are Hindi or English or Regional.

- **ATYPE** - The admission type which may be through Uttar Pradesh State Entrance Examination (UPSEE) or direct admission through University procedure.

- **RET** – Retention. Whether the student continue or not after one year. Possible values are 1 if student continues study and 0 if student dropped the study after one year.

*A. Implementation of Mining Model*

Weka is open source software that implements a large collection of machine leaning algorithms and is widely used in data mining applications. From the above data, ret.arff file was created. This file was loaded into WEKA explorer. The classify panel enables the user to apply classification and regression algorithms to the resulting dataset, to estimate the accuracy of the resulting predictive model, and to visualize erroneous predictions, or the model itself. There are 16 decision tree algorithms like ID3, J48, ADT etc. implemented in WEKA. The algorithm used for classification is ID3, C4.5 and ADT. Under the "Test options", the 10-fold cross-validation is selected as our evaluation approach. Since there is no separate evaluation data set, this is necessary to get a reasonable idea of accuracy of the generated model. The model is generated in the form of decision tree. These predictive models provide ways to predict whether a new student will continue to enroll or not after one year.

*B. Results and Discussion*

The three decision trees as examples of predictive models obtained from the retention data set by three machine learning algorithms: the ID3 decision tree algorithm, the C4.5 decision tree algorithm and the alternative decision tree (ADT) algorithm. For example, consider a new case with a student having graduation with B.Sc. (GS= B.Sc. with Math), and graduation grade is First (GOG = First), category is General (Cat = General) and get admission from UPSEE (ATYPE = UPSEE). For both the ID3 decision tree and the C4.5 decision tree, we need to start from the root to find a unique path leading to a prediction leaf node. In both trees, we find a unique path of length 1 immediately leading us from the root to a leaf node labeled 1, predicting continued enrollment the next year.

```
: -0.506
|  (1)GSS = A: -1.261
|  (1)GSS != A: 0.389
|  (2)MED = Hindi: 0.263
|  |  (4)GOG = Second: 0.433
|  |  |  (7)GMSS = A: -0.512
|  |  |  (7)GMSS != A: 0.476
|  |  |  |  (10)GSS = A: -0.372
|  |  |  |  (10)GSS != A: 0.564
|  |  (4)GOG != Second: -0.485
|  |  |  (8)GSS = C: 0.438
|  |  |  (8)GSS != C: -0.614
|  (2)MED = English: -0.372
|  |  (3)GMSS = C: -0.756
|  |  (3)GMSS != C: 0.23
|  |  |  (5)GS = B.A.  without maths: 0.507
|  |  |  |  (9)Cat = OBC: -0.412
|  |  |  |  (9)Cat != OBC: 0.603
|  |  |  (5)GS != B.A.  without maths: -0.384
|  |  |  |  (6)GSS = 0: 0.533
|  |  |  |  (6)GSS != 0: -0.653
```

Figure 1: AD Tree.

On the other hand, for the alternative decision tree (ADT tree) as shown in fig. 1, we may have multiple paths from the root to the leaves that are consistent with data and we need to sum up all the numbers appearing on these paths to see whether it is positive or negative. In this particular case, we find three paths leading from the root to leaves. Summing up all the numerical numbers appearing on these paths, we have a positive value 0.483+0.15-0.218+0.125+0.036=0.576, and that leads to the prediction of continued enrolment too. These decision trees also provide interesting insights into hidden patterns in the student retention data set. For example, both the ADT tree and the C4.5 decision tree show that Graduate Stream (GS) is a very relevant factor.

The Table II shows the accuracy of ID3, C4.5 and CART algorithms for classification applied on the above data sets using 10-fold cross validation is observed as follows:

TABLE II: CLASSIFIERS ACCURACY

| Algorithm | Correctly Classified Instances | Incorrectly Classified Instances |
|---|---|---|
| ID3 | 72.093% | 11.627% |
| C4.5 | 74.416% | 25.581 % |
| ADT | 72.093% | 27.907% |

Table II shows that a C4.5 technique has highest accuracy of 74.416% compared to other methods. ID3 and ADT algorithms also showed an acceptable level of accuracy.

The Table III shows the time complexity in seconds of various classifiers to build the model for training data.

TABLE III: EXECUTION TIME TO BUILD THE MODEL

| Algorithm | Execution Time (Sec) |
|---|---|
| ID3 | 0.12 |
| C4.5 | 0.08 |
| ADT | 0.06 |

Table IV below shows the three machine learning algorithms that produce predictive models with the best precision values in our experiments, together with the corresponding recall values. For these algorithms, the best precision values (ranging from around 68.2% to 82.8%) are almost all accomplished when learning from the data set, with recall values ranging from 6.4% to 11.4%.

The alternative decision tree (ADT) learning algorithm is the best precision performer we have seen so far, capable of reaching a precision rate of 82.8% and a recall rate of 11.4% without a sign of over-fitting. In other words, given a collection of 1000 new first year students with around 250 would-be drop-out cases embedded in the list (assuming a drop-out rate of 25%), the ADT tree algorithm is likely to produce a list of around 37 students and among them about 31 are actual would-be drop-out cases.

TABLE IV: CLASSIFIERS ACCURACY

| Algorithm | Precision values | Recall values |
|---|---|---|
| ID3 | 68.2% | 06..4% |
| C4.5 | 70.4% | 09.6 % |
| ADT | 82.8% | 11.4% |

## V. CONCLUSIONS

One of the data mining techniques i.e., classification is an interesting topic to the researchers as it is accurately and efficiently classifies the data for knowledge discovery. Decision trees are so popular because they produce classification rules that are easy to interpret than other classification methods. Frequently used decision tree classifiers are studied and the experiments are conducted to find the best classifier for retention data to predict the student's drop-out possibility. Machine learning algorithms such as the alternative decision tree (ADT) learning algorithm can learn effective predictive models from the student retention data accumulated from the previous years. The empirical results show that we can produce short but accurate prediction list for the student retention purpose by applying the predictive models to the records of incoming new students. This study will also work to identify those students which needed special attention to reduce drop-out rate.

AUTHORS PROFILE

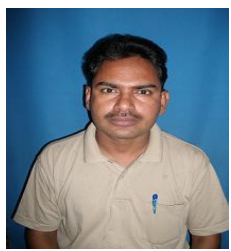

Surjeet Kumar Yadav received his M.Sc. (Computer Science) from Dr. Baba Sahed Marathwada University, Aurangabad, Maharastra, India (1998). At present, he is working as Sr. Lecturer at Department of Computer Applications, VBS Purvanchal Uniersity, Jaunpur. He is an active member of CSI and National Science Congress. He is currently doing research in Data Mining and Knowledge Discovery.

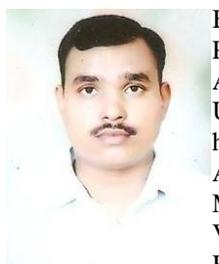

Brijesh Kumar Bhardwaj is Assistant Professor in the Department of Computer Applications, Dr. R. M. L. Avadh University Faizabad India. He obtained his M.C.A degree from Dr. R. M. L. Avadh University Faizabad (2003) and M.Phil. in Computer Applications from Vinayaka mission University, Tamilnadu. He is currently doing research in Data Mining and Knowledge Discovery. He has published two international papers.

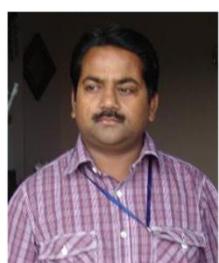

Saurabh Pal received his M.Sc. (Computer Science) from Allahabad University, UP, India (1996) and obtained his Ph.D. degree from the Dr. R. M. L. Awadh University, Faizabad (2002). He then joined the Dept. of Computer Applications, VBS Purvanchal University, Jaunpur as Lecturer. At present, he is working as Head and Sr. Lecturer at Department of Computer Applications.

Saurabh Pal has authored more than 25 research papers in international/national Conference/journals and also guides research scholars in Computer Science/Applications. He is an active member of CSI, Society of Statistics and Computer Applications and working as reviewer for more than 15 international journals. His research interests include Image Processing, Data Mining, Grid Computing and Artificial Intelligence.